\documentclass[letterpaper, 10 pt, conference]{ieeeconf}  

\IEEEoverridecommandlockouts                              

\usepackage{epsfig} 
\usepackage{cite}
\usepackage{times} 
\usepackage{amsmath} 
\usepackage{nicefrac}
\usepackage{amssymb}  
\usepackage{graphicx}
\usepackage{mathrsfs}   
\usepackage{caption}
\usepackage{subcaption}
\usepackage{url}
\usepackage{color}

\usepackage{amsthm}

\usepackage[linesnumbered,ruled,vlined]{algorithm2e}
\usepackage{amsthm}
\usepackage[english]{babel}
\usepackage{multirow}
\usepackage{booktabs,caption}
\usepackage[flushleft]{threeparttable}
\usepackage{diagbox}
\usepackage{graphicx}
\usepackage{svg}

\graphicspath{ {./figures/} }

\title{\LARGE{ \bf{
Joint Communication and Motion Planning for Cobots 
}} \\  Extended Version
}

\author{Mehdi Dadvar$^{1}$, Keyvan Majd$^{1}$, Elena Oikonomou$^{1}$, Georgios Fainekos$^{1}$, and Siddharth Srivastava$^{1}$ 
\thanks{This work was supported in part by the NSF under grants IIP-1361926, IIS-1909370, IIS-1844325, OIA-1936997 and the NSF I/UCRC Center for Embedded Systems.}%
\thanks{$^{1}$ The authors are with the School of Computing and Augmented Intelligence, Arizona State University, Tempe, AZ, USA.
        {\tt\small \{mdadvar,majd,	
e.oikonomou,fainekos,siddharths\}@asu.edu}}%
}

\SetKwInput{KwInput}{Input}               
\SetKwInput{KwOutput}{Output}

\theoremstyle{definition}
\newtheorem{definition}{Definition}
\newtheorem{Lemma}{Lemma}
\newtheorem{Assumption}{Assumption}

\newtheorem{theorem}{Theorem}

\newtheorem{proposition}{Proposition}
\begin{document}

\maketitle
\thispagestyle{empty}
\pagestyle{empty}


\begin{abstract}
 The increasing deployment of robots in co-working scenarios with humans has revealed complex safety and efficiency challenges in the computation of the robot behavior. Movement among humans is one of the most fundamental \textemdash and yet critical\textemdash problems in this frontier. While several approaches have addressed this problem from a purely navigational point of view, the absence of a unified paradigm for communicating with humans limits their ability to prevent deadlocks and compute feasible solutions.  This paper presents a joint communication and motion planning framework that selects from an arbitrary input set of robot's communication signals while computing robot motion plans. It models a human co-worker's imperfect perception of these communications using a noisy sensor model and facilitates the specification of a variety of social/workplace compliance priorities with a flexible cost function. Theoretical results and simulator-based empirical evaluations show that our approach efficiently computes motion plans and communication strategies that reduce conflicts between agents and resolve potential deadlocks.
\end{abstract}

\section{Introduction}
\begin{figure}[t]
      \centering
      \includegraphics[scale=0.25]{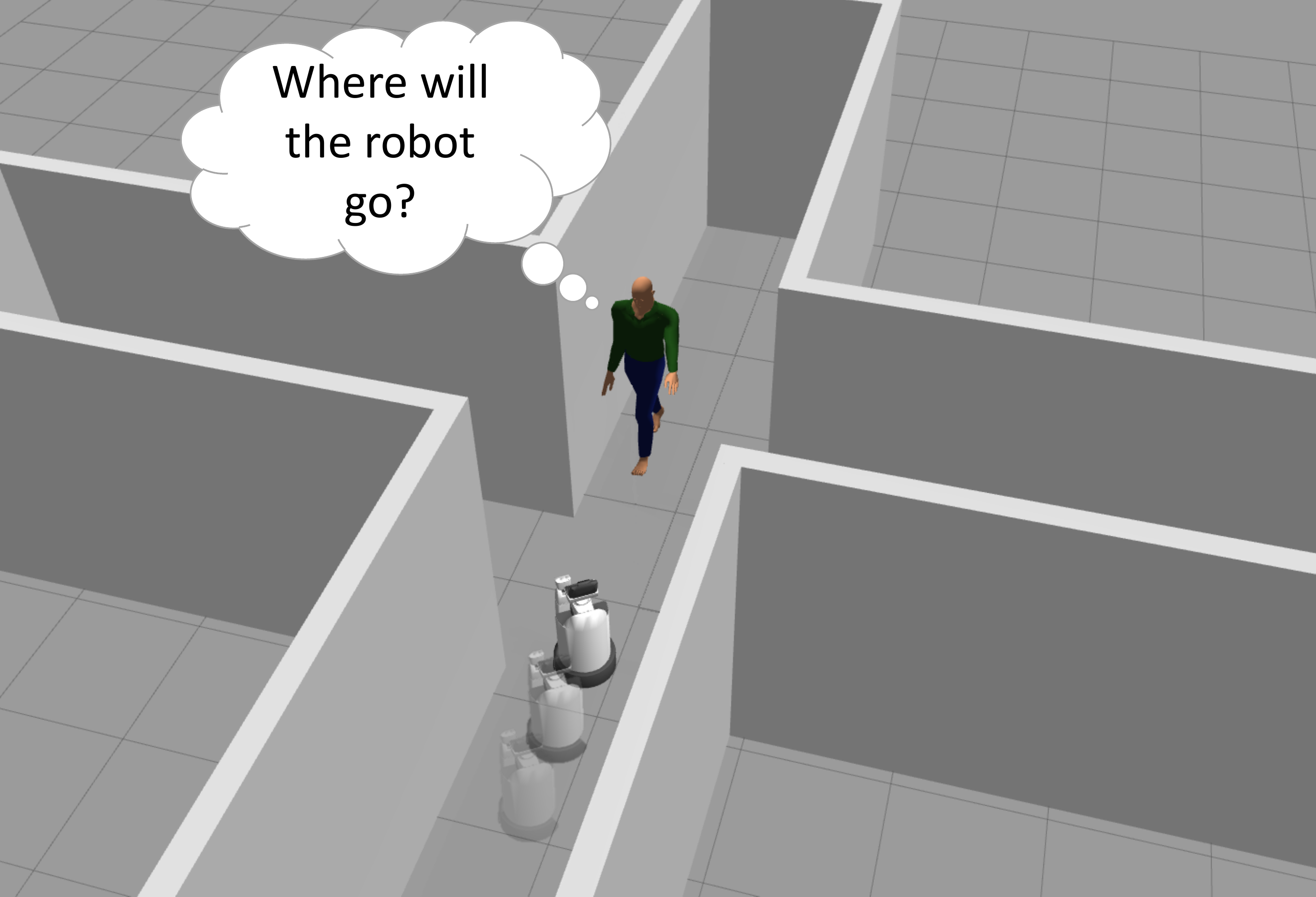}
      \caption{An example of a social navigation scenario in a confined environment where the robot's movement can not reveal any information about its future intentions.}
      \label{fig:example problem}
   \end{figure}

Technological breakthroughs of the past decade have led to
increasingly common human-robot co-working
environments~\cite{cheng2018autonomous}.  Navigating among humans is
an imperative task that most cobots, ranging from industrial to
service robots, are expected to perform safely and
efficiently. Although 
motion planning for autonomous robots has been studied from multiple
perspectives~\cite{chen2017socially,kuderer2012feature,trautman2013robot},
these approaches focus on movement actions and do not address the
problem using communication to resolve situations that require
extensive human-robot interaction.
The objective of this paper is to develop a unified paradigm for
computing movement and communication strategies that improve
efficiency and reduce movement conflicts in co-working scenarios (see Fig.\,\ref{fig:example problem}).

Although the problem of integrated task and motion planning has received significant research attention \cite{kaelbling2010hierarchical,garrett2020pddlstream,srivastava2014combined,shah2020anytime, dantam2018incremental,dantam2018task} the integration of these deliberative processes with communication has not been studied sufficiently. Prior work on this topic includes extensions to sampling based motion
planning paradigms that model
pedestrians as moving obstacles~\cite{sucan2012open,kuffner2000rrt}. While these extensions provide
valuable enhancements of well-known and efficient algorithms, they
view humans as impervious entities and have limited applicability in
co-working scenarios where both the human and the robot need to adjust their
behavior to allow feasible
solutions. 
On the other hand, there are approaches that employ disjoint
prediction models to establish simple interactions with humans to
generate safer and more risk-aware motion plans~\cite{nishimura2020risk}. Since these approaches neglect the effect of
the robot's motion on the human's behavior, they suffer from \emph{the robot freezing problem} where the robot cannot find any safe
solution. To address this limitation,
\emph{socially compliant} methods consider potential
human-robot cooperation via learning and planning techniques to produce legible plans or plans subject
to stipulations on the information divulged during plan
execution \cite{trautman2015robot,kulkarni2019unified,zhang2018finding}.
\cite{kretzschmar2016socially} employs inverse reinforcement learning
(IRL) to learn interactive models of pedestrians in the environment
for social compliant path planning. Further,~\cite{kivrak2021social}
presents a social navigation framework that adapts the social force model
(SFM) to generate human-friendly collision-free motion plans in
unknown environments.
Although these approaches model the effect of the robot's movement on
the humans' behavior for legible motion planning,
relying purely on motion actions, taxonomically known as
implicit communication~\cite{knepper2017implicit}, could be misleading
for the human~\cite{habibian2021here} and may lead
to deadlocks in confined environments.

Clearly,
employing explicit communication~\cite{baraka2018mobile}
coupled with the robot's movements would
enrich the human-robot interaction. ~\cite{che2020efficient} uses IRL to model the effects of both explicit and implicit actions of
the robot on the human's behavior. Further, a robot planner relies on
this model to produce communicative actions to maximize the robot's
clarity. Since this method assumes predefined behavior modes for the robot and human (robot priority and human priority), the solution always impels one agent to slow down, which
degrades the planning effectiveness.

In contrast, we formalize a unified deliberative communication planning
problem that addresses the joint problem of computing the robot's
communication strategy and movements while taking into account the
human's imperfect perception about the robot and its communications
(Sec.\,\ref{sec:methodology}).  We use a noisy communication model to
estimate the results of robot's communications on the human's belief
of the robot's possible locations. In contrast to the human prediction
framework in~\cite{che2020efficient}, which requires the robot's
future trajectories (the need for socially compliant planning
illustrates the difficulty of obtaining such inputs), our approach supports arbitrary human movement prediction models that can predict human
behaviors given a set of possible obstacles. Our solution paradigm
derives estimates of the human's belief on the robot's positions to
compute robot communication and movement plans
(Sec.\,\ref{sec:communication planner})). This is done using a hierarchical search process with a socially compliant motion planner Control Barrier
Function enabled Time-Based RRT (CBF-TB-RRT)~\cite{majd2021safe}
(Sec.\,\ref{sec:motionplanner}). Theoretical results and extensive
simulations on various test environments show that this approach
efficiently avoids deadlocks and computes mutually efficient solutions
without requiring preset behavior modes.

\section{Preliminaries}
\subsection{Control Barrier Function (CBF)}
Assume that the robot $R$ is following a nonlinear control affine dynamics as
\begin{align}
\label{eq: sys_dyn1}
 \dot{\mathbf{s}}_r =\mathbf{f}_r(\mathbf{s}_r)+\mathbf{g}_r(\mathbf{s}_r)\mathbf{a}_r,
\end{align}
where $\mathbf{s}_r\in \mathcal{S}_R \subseteq \mathbb{R}^{n}$ denotes the state of $R$,  $\mathbf{a}_r\in \mathcal{A}_R \subseteq \mathbb{R}^{m}$ is the control input, and $\mathbf{f}_r:\mathbb{R}^{n} \rightarrow \mathbb{R}^{n}$ and $\mathbf{g}_r:\mathbb{R}^{n} \rightarrow \mathbb{R}^{n \times m}$ are locally Lipschitz functions. \

A function $\alpha: \mathbb{R} \rightarrow \mathbb{R}$ is an extended class $\mathcal{K}$ function iff it is strictly increasing and $\alpha(0) = 0$ \cite{ames2019control}.
A set $\mathcal{C} \subseteq \mathbb{R}^{n}$ is forward invariant w.r.t the system (\ref{eq: sys_dyn1}) iff  for every initial state $\mathbf{s}^0_r \in \mathcal{C}$, its solution satisfies $\mathbf{s}^t_r\in \mathcal{C}$ for all $t \geq 0$ \cite{blanchini1999set}.

\begin{definition}[Control Barrier Function~\cite{ames2019control}]
A continuously differentiable function $B(\mathbf{s}_r)$ is a Control Barrier Function (CBF) for the system (\ref{eq: sys_dyn1}), if there exists a class $\mathcal{K}$ function $\alpha$ s.t. $\forall \mathbf{s}_r\in \mathcal{C}$ :
\begin{equation}\label{eq: CBF}
\sup_{a_r\in \mathcal{A}_R}\big(L_{f_r} B(\mathbf{s}_r) +L_{g_r} B(\mathbf{s}_r) \mathbf{a}_r +\alpha(B(\mathbf{s}_r))\big)\geq 0 
\end{equation}
where $L_{f_r} B(\mathbf{s}_r) = \frac{\partial B}{\partial \mathbf{s}_r}^\top f_r(\mathbf{s}_r), L_{g_r} B(\mathbf{s}_r)= \frac{\partial B}{\partial \mathbf{s}_r}^\top g_r(\mathbf{s}_r)$ are the first order Lie derivatives of the system.
\end{definition}

Any Lipschitz continuous controller $\mathbf{a}_r \in K_{cbf}(\mathbf{s}_r) = \{\mathbf{a}_r\in \mathcal{A}_R\;|\;L_f B(\mathbf{s}_r) +L_{g_r} B(\mathbf{s}_r) \mathbf{a}_r +\alpha(B(\mathbf{s}_r))\geq 0\}$ results in a forward invariant set $\mathcal{C}$ for the system (\ref{eq: sys_dyn1}).

\subsection{Control Barrier Function Enabled Time-Based Rapidly-exploring Random Tree (CBF-TB-RRT)}
CBF-TB-RRT, proposed in~\cite{majd2021safe}, provides a probabilistic safety guaranteed solution in real-time to the start-to-goal motion planning problem. At each time step, given a probabilistic trajectory of dynamic agents, this method extracts ellipsoidal reachable sets for the agents for a given time horizon with a bounded probability. This method extends time-based RRT (each node of TB-RRT denotes a specific state in a specific time), proposed in~\cite{sintov2014time}, in conjunction with CBFs to generate path segments for $R$ (Eq. (\ref{eq: sys_dyn1})) that avoid the agents' reachable sets while moving toward goal. If the probability distribution over the dynamic agents' future trajectory for a given finite time horizon is accurate, the generated control by CBF-TB-RRT guarantees that the probability of collision at each time step is bounded.

\section{Deliberative Communication Planning} \label{problem}
We formulate the deliberative communication planning problem $\mathcal{P_{DC}}$ as the problem of jointly computing communication signals with corresponding feasible motion plans for $R$ in a social navigation scenario. As a starting point, we focus on settings with a single robot and a single human $H$. In such problems, $R$'s actions $\mathcal{A}$ include communication as well as movement actions. In order to model realistic scenarios, we use potentially noisy models of $H$'s movement ($T_H$) and of $H$'s sensing ($O$) of $R$'s communications. We use these models to  evaluate possible courses of action while computing efficient, collision-free communication and movement plans for $R$. 

Intuitively, $T_H$ maps the current state of $H$ and $H$'s belief about the possible positions of $R$ at the next planning cycle to possible motion plans for $H$. We model $H$'s sensor model $O$ as a variation of the standard noisy sensor paradigm used in planning under partial observability. $O$ relates $H$'s current state, $R$'s communication action and $R$'s intended next state to the observation signal that $H$ receives. In this formulation, $H$ \emph{need not know $R$'s current/intended states nor the exact communication that it executed} -- $H$ only receives an observation signal. Such sensor models are very general: they can capture a variety of scenarios ranging from perfect communication to imperfect communication settings where $H$ may not have a perfect understanding or observation of $R$'s communications and may conflate $R$'s communication actions with each other.

\begin{definition} \label{def:problem}

A deliberative communication planning problem is a tuple $\mathcal{P_{DC}} =\langle  \mathcal{S}, s^0, \mathcal{A}, T, \mathcal{G},O, J\rangle$, where:
\begin {itemize}
\item $\mathcal{S} = \mathcal{S}_R\times \mathcal{S}_H$ is the set of states consisting of  $R$'s and $H$'s states, respectively.
\item $s^0 = s^0_r\times s^0_h$ are the initial states of $R$ and $H$, respectively, where $s^0_r\in \mathcal{S}_R$ and $s^0_h\in \mathcal{S}_H$.
\item $\mathcal{A}$ is the set of $R$'s actions defined as $\mathcal{A} = \mathcal{A}_c \cup \mathcal{A}_m$, where $\mathcal{A}_c$ is a set of  communication signals that includes the null communication, and $\mathcal{A}_m$ is the implicit uncountable set of $R$'s feasible motion plans. Each feasible motion plan $\pi_R\in \mathcal{A}_m$ is a continuous function  $\pi_R:[0,1]\rightarrow \mathcal{S}_R$ where $\pi_R(0) = s^0_r$ and $\pi_R(1) \in \mathcal{S}_R$. 
\item $\mathcal{G}=\langle \mathcal{G}_R, \mathcal{G}_H\rangle$ is the goal pair where  $\mathcal{G}_R\subseteq \mathcal{S}_R$ is $R$'s goal set and  $\mathcal{G}_H\subseteq \mathcal{S}_H$ is $H$'s goal set.
\item $T = \langle T_R, T_H \rangle$ constitutes transition/movement models of both agents where $T_R$ is $R$'s transition function defined as $T_R: \mathcal{S}_R \times \mathcal{A}_m \rightarrow \mathcal{S}_R$, and  $T_H: \mathcal{S}_H \times \mathcal{G}_H \times \mathcal{B}^{R'}_H \rightarrow 2^{\Pi_H}$ denotes $H$'s movement model where $\mathcal{B}^{R'}_H$ is the set of possible beliefs over the state of $R$ at the next planning cycle and $\Pi_H$ is the set of feasible $H$ movement plans within $\mathcal{S}_H$. $T_H$ 
may be available as a simulator that yields a sample of the possible $H$ plans.
\item $O$ is $H$'s sensor model defined as $O: \mathcal{S}_H\times \mathcal{A}_c \times \mathcal{S}_R \rightarrow \Omega $, where $\Omega$ denotes $H$'s observation. Situations where $H$ cannot perfectly understand or observe $R$'s communication can be modeled by  mapping multiple tuples $\langle \mathbf{s}_h, a_c, \mathbf{s}_r\rangle$ to the same $\omega\in \Omega$, where $a_c \in \mathcal{A}_c$.
\item $J: \mathcal{S}_H\times \mathcal{S}_R\times \mathcal{A} \rightarrow \mathbb{R}$ is a utility function denoting the value of a joint $H$-$R$ state and a communication-motion action. In practice, we express $J$ as a cost function.   
\end {itemize}
\end{definition}

A solution to $\mathcal{P_{DC}}$ is a sequence of communication actions and motion plans that satisfy $\mathcal{G}_R$, and is defined as follow.
\begin{definition}
 A solution to the deliberative communication planning problem $\mathcal{P_{DC}} =\langle  \mathcal{S}, s^0, \mathcal{A}, T, \mathcal{G}, O, J\rangle$ is a finite sequence of communication and movement actions: $\Psi = \langle (a_c^1, \pi_R^1), (a_c^2, \pi_R^2), \cdots,(a_c^q,\pi_R^q)  \rangle$, where $a_c^i \in \mathcal{A}_c$, $\pi_R^i \in \mathcal{A}_m$, $\pi_R^{1}(0)=s_r^0$, $\pi_R^i(1)=\pi_R^{i+1}(0)$, and $\pi_R^q(1)\in \mathcal{G}_R$ for $i=1,\cdots,q$. 
 
\end{definition}

\section{Methodology} \label{sec:methodology}
\subsection {Overview} \label{sec:overview}
In the proposed paradigm of joint communication and motion planning, a motion planner (MP) returns a set of feasible and collision-free motion plans $\Pi_R \in \mathcal{A}_m$ considering the goal set $\mathcal{G}$.
Accordingly, a communication planner (CP) uses a search tree to select a combination of a communication action and a motion plan at each planning cycle that minimizes $J$. 
Each node of this search tree is defined by $\langle \mathbf{s}, a_c, \pi_R\rangle$ where  $\mathbf{s}\in \mathcal{S}$, $a_c \in \mathcal{A}_c$ and $\pi_R \in \Pi_R$. $a_c$ denotes the communication action being considered at this node while $\pi_R$ denotes one of the plans returned by MP.

\begin{figure}[t]
      \centering
      \includegraphics[scale=0.42]{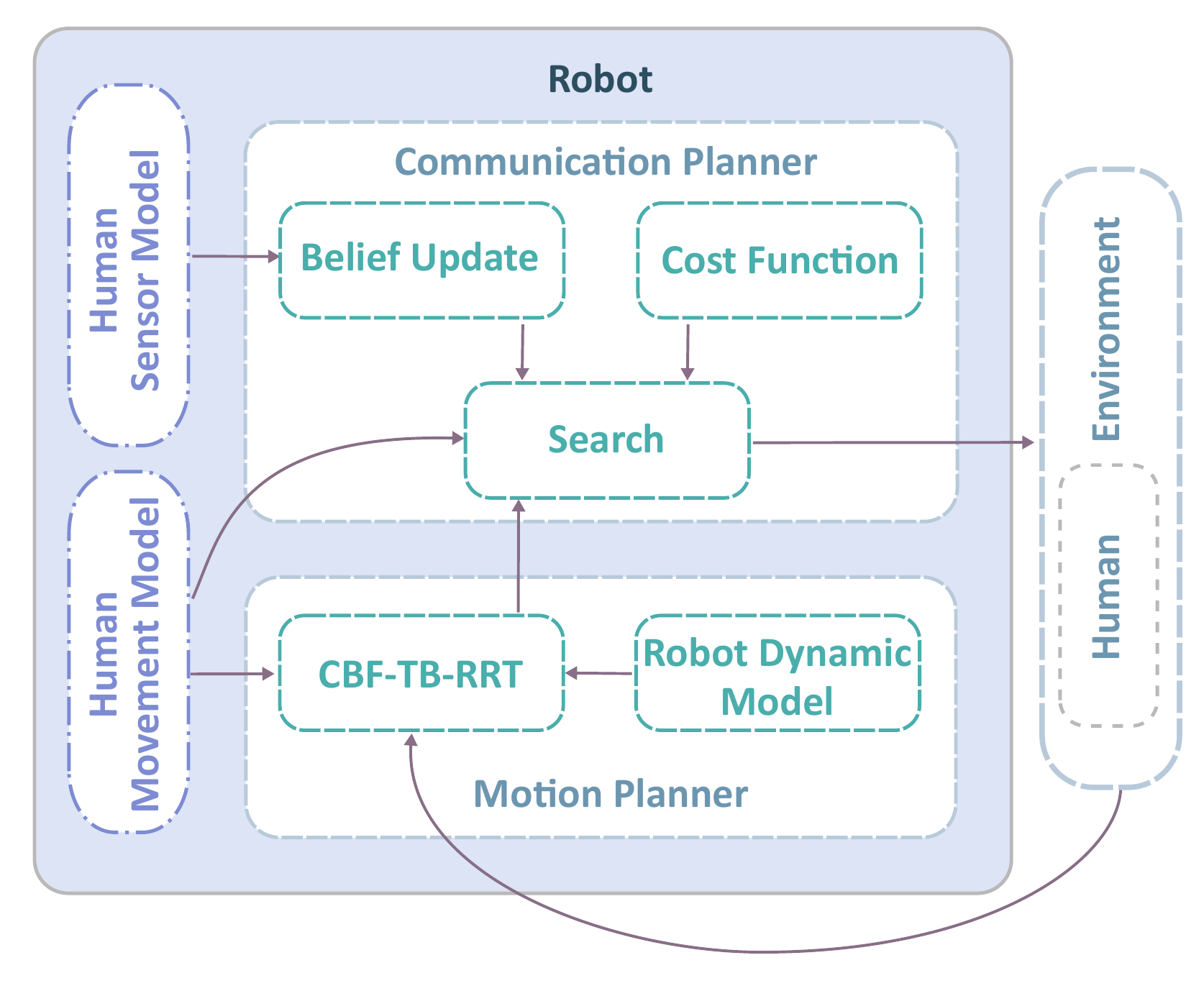}
      \caption{An overview of our approach.}
      \label{fig:blockdiagram}
   \end{figure}

Fig.\,\ref{fig:blockdiagram} illustrates the mechanism by which MP and CP interact. MP utilizes CBF-TB-RRT with $H$'s movement model and  $R$'s dynamic model to produce a finite set of feasible motion plans $\Pi_R \subset \mathcal{A}_m$ (Sec.\,\ref{sec:motionplanner}). Starting with a node representing the current state, CP creates a successor node for each combination of a feasible plan in $\Pi_R$ and a communication action from $\mathcal{A}_c$. For each such combination, it uses a belief update process to compute and store an estimate of $H$'s next belief if $R$ were to use the corresponding communication action. At each planning cycle, CP selects a node of tree that minimizes $J$ (CP is described in Sec.\,\ref{sec:communication planner}). An important property of this approach is that our solution algorithms are independent of the choice of $R$, the environment, and $H$'s movement and sensor models. 

\subsection {CBF-based TB-RRT as MP} \label{sec:motionplanner}
We obtain a set $\Pi_R$ of diverse plans in Alg. \ref{alg: select_plan_min} by employing CBF-TB-RRT~\cite{majd2021safe} as MP. 
We modified the original CBF-TB-RRT method to better serve our hierarchical framework as follows. First, the set of possible future trajectories for $H$ can either be given by a stochastic $T_H$ or by a deterministic $T_H$ with an $\varepsilon$ tube around the predicted trajectory. We denote this region by $\mathcal{S}_R^{unsafe}$. MP maintains a continually updated estimate of $R$'s safe states, $\mathcal{S}^{safe}_R=\mathcal{S}_R\setminus \mathcal{S}^{unsafe}_R$, where $\mathcal{S}_R^{safe}$ would be collision-free with respect to the predicted trajectories of $H$. Second, the original CBF-TB-RRT expands a tree for a finite time horizon and just apply the control for the first time-step at each planning cycle. In contrast, here, we let $R$ to execute the full returned partial plan. Finally, instead of selecting one plan to execute, we select a set of $p$ plans $\Pi_R\subseteq \mathcal{A}_m$ of $\Bar{\pi}_{R,j}: [t_0,t_j] \rightarrow \mathcal{S}^{safe}_R$ for $j=\{1,2,\cdots,p\}$. Here, each plan $\Bar{\pi}_{R,j}$ represents a path segment from the initial vertex $\nu_0$ at location $\Bar{\pi}^{t_0}_R$ in time $t_0$ to another vertex $\nu_j$ at location $\Bar{\pi}^{t_j}_R$ in time $t_j$. Assuming $c_j$ to be the cost of vertex $\nu_j$ in the set of all expanded RRT vertices $\mathcal{V}$, we minimize the following cost to select $p$ diverse plans $\Bar{\pi}_{R,j}$ with the minimum costs $c_j$ for $j\in \{1,2,\cdots,p\}$,
\begin{align}
       & \underset{\mathbf{r}}{\text{min}}
       & & J_d = \sum^{\lvert\mathcal{V}\rvert}_{i=0}\frac{w_c r_i c_i}{w_d\sum^{\lvert\mathcal{V}\rvert}_{j=1, i\neq j}r_j d_{ij}}, \nonumber\\
       & \label{eq: select_plan_min} \text{s.t.} & &   \sum_{i=0}^{\lvert\mathcal{V}\rvert} r_i = p, \\
       & & & r_i\in \{0,1\}, & & & \text{for }i=0,\cdots,\lvert \mathcal{V}\rvert, \nonumber
\end{align}
where $w_c$ and $w_d$ are the numerator and denominator weights, respectively, $d_{ij}$ is the Euclidean distance between vertices $i$ and $j$, and $\mathbf{r}$ is a vector of binary values $r_i$ for $i=1,\cdots,\lvert \mathcal{V}\rvert$, that determines the selected plans (vertices). Given the expanded RRT at each planning cycle, we minimize (\ref{eq: select_plan_min}) using Alg.\,\ref{alg: select_plan_min} to find $p$ diverse $\Bar{\pi}_R$ plans. 

\begin{algorithm}[h]
\DontPrintSemicolon
\SetAlgoLined
\KwInput{$\mathcal{V}$ and $p$ }
\KwOutput{$\Pi_R$}
$\mathcal{P}\leftarrow$ Randomly select $p$ vertices from $\mathcal{V}$\\
$\textsc{Opt\_Cost}\leftarrow$ Calculate the cost $J_d$ for the vertices in $\mathcal{P}$  \\
\While{CONVERGE}{
\For{$\nu\in \mathcal{V}\setminus\mathcal{P}$}{\label{alg: select_plan_min-line4}
Calculate the cost $J_d$ for all $p$-combinations of $\mathcal{P}\cup \{\nu\}$ and update $\textsc{Opt\_Cost}$ and $\mathcal{P}$ with the minimum cost combination
}
}
$\Pi_R\leftarrow$ Extract the path segments $\Bar{\pi}_R$ from $\nu_0$ to each $p$ vertex in $\mathcal{P}$ 
\caption{RRT Plan $\Pi_R$ Generation (MP)}
\label{alg: select_plan_min}
\end{algorithm}

\begin{proposition}
Given that RRT includes a finite set of vertices and $J_d\geq 0$, Alg.\,\ref{alg: select_plan_min} terminates in a finite time.
\end{proposition}

\begin{Assumption}\label{assum: human-pred}
The future human motion remains within the unsafe region $\mathcal{S}^{unsafe}_R$ predicted by $T_H$.
\end{Assumption}
\begin{Lemma} \label{lem:cbf safe}
Following Assn. \ref{assum: human-pred}, all generated path segments $\Bar{\pi}_{R,j}$ for $j=1,\cdots,\lvert\mathcal{V}\rvert$ by CBF-TB-RRT are guaranteed to remain in $\mathcal{S}^{safe}_R$ if $\Bar{\pi}^{t_0}_R\in \mathcal{S}^{safe}_R$.
\end{Lemma}
\begin{proof}
This proof is immediate following \cite[Prop. 1]{majd2021safe}.
\end{proof}

\subsection {Communication Planner Module (CP)} \label{sec:communication planner}
As discussed in Sec.\,\ref{sec:overview}, CP builds a search tree to select an optimal combination of communication action and motion plan. Recall that each node in the search tree consists of a state $\mathbf{s}$, a communication action $a_c$ and a motion plan $\pi_R$. Here, $\pi_R$ denotes the discretization of the continuous-time path segment $\Bar{\pi}_R$ given by MP. We use a belief-space formulation to represent the set of locations where $H$ might expect $R$ to be at the next planning cycle $k+1$. Thus, the set of all possible beliefs of $H$, is the power set of $\mathcal{S}_R$. However, in practice $H$ needs to keep track of only a subset of possible locations, in a small neighbourhood around $H$.

\begin{definition}
A \emph{\textbf{$\delta$-local neighborhood}} of $H$ is a subset $\mathcal{L}\subseteq \mathcal{S}_R$ s.t. the Euclidean distance from $S_H$ $d(\mathbf{s}_{xyz}, \mathcal{S}_H)$ of $R$'s base coordinates $\mathbf{s}_{xyz}$ in state $\mathbf{s}$ is less than $\delta$ $\forall S \in \mathcal{L}$.
\end{definition}

We maintain a bounded, discretized set of regions to approximate $H$'s belief about $R$'s presence in their $\delta$-local neighborhood. Let $\mathcal{L}_H$ be the  set of these discretized zones $\{l_1,\ldots,  l_{\ell}\}$. Collectively these regions can represent neighborhoods in domain-specific configurations (e.g., an $H$-centered forward-biased cone or a rectangular region around $H$ with discretized cells). Given a state $(\mathbf{s}_R, \mathbf{s}_H)\in \mathcal{S}$ we use $\mathbf{s}_R\in l_i(\mathbf{s}_H)$ to express that when $R$'s state is  $\mathbf{s}_R$ and $H$'s state is $\mathbf{s}_H$, $R$ will be in the region $l_i$ in $H$'s local neighborhood.
In this notation, $H$'s belief is a Boolean vector of dimension $\lvert l_H\rvert$, so that $b_i=1$ in a belief $\mathbf{b}$ represents a belief that $\mathbf{s}_R\in l_i(\mathbf{s}_H)$ is possible at the next time step.

Given a starting belief $\mathbf{b}_k$ and an observation symbol $\omega_k$, we
can invert the sensor model and the transition function to derive a
logical filtering based belief update equation for computing $\mathbf{b}_{k+1}$
as follows.  Let $\varphi_1 (\mathbf{s}_R^{k+1}, i)$ state that $R$ at
$\mathbf{s}_R^{k+1}$ would be in $H$'s $i^{th}$ neighborhood zone, i.e.,
$\mathbf{s}_R^{k+1}\in l_i(\mathbf{s}_H^{k})$; $\varphi_2(\mathbf{s}_R^k,j)$ state that
$b_j^k$ was 1 with $R$ at $\mathbf{s}_R$, i.e., $ b^k_j=1 \land \mathbf{s}_R^k\in
l_j(\mathbf{s}_H^{k-1})$; $\varphi_3(\mathbf{s}_R^k, \mathbf{s}_R^{k+1})$ state that $R$ can move
from $\mathbf{s}_R^{k}$ to $\mathbf{s}_R^{k+1}$, i.e., $\exists a_m \in \mathcal{A}_m, T_R(\mathbf{s}_R^k,a_m) =\mathbf{s}_R^{k+1} $; and $\varphi_4(\mathbf{s}_H^k,\omega, \mathbf{s}_R^{k+1})$ state that $R$ may have executed a communication action $a_c$ that resulted in observation $\omega$, i.e., $\exists a_c\in \mathcal{A}_c, o(\mathbf{s}_H^k, a_c, \mathbf{s}_R^{k+1})=\omega$.  Inverting the sensor model and the transition function gives us $b_i^{k+1}=1 \emph{iff}$ $\exists \mathbf{s}_R^{k},
\mathbf{s}_R^{k+1}\in \mathcal{S}_R; j \in [1, \ell]:$ $\varphi_1(\mathbf{s}_R^{k+1}, i) \land \varphi_2(\mathbf{s}_R^k, j)$ $\land \varphi_3(\mathbf{s}_R^k, a_m, \mathbf{s}_R^{k+1})\land\varphi_4(\mathbf{s}_H^k,\omega, \mathbf{s}_R^{k+1})$. CP uses this
expression to compute $R$'s estimate of $H$'s belief $\mathbf{b}^{k+1}$ given a belief $\mathbf{b}^k$ at the parent node and the observation  $\omega$ that $H$ would receive as a result of the communication action being considered at that node. We use $\mathbf{b}(n)$ to denote this belief for  node $n$.

CP uses a cost function $J$ to evaluate a node $n=\langle \mathbf{s}, a_c, \pi_R\rangle$ in the search tree. Intuitively, $J$ needs to consider $H$ and $R$'s future paths $\Gamma_H$ and $\Gamma_R$, respectively. $\tilde{\Gamma}_R(n)$ is an estimate for $\Gamma_R$ based on $\pi_R$. 
However, we do not have an accurate future path for $H$ and we use $\mathbf{b}(n)$ and the human movement model $T_H$ to obtain an estimate $\tilde{\Gamma}_H(n)$. We omit the node argument unless required for clarity.

For computational efficiency, we discretize $\Gamma_R$ and $\Gamma_H$ as sequences of waypoints: $\Gamma_R=\{ \gamma_R^i \}_{i=1}^{i_{max}}$ and $\Gamma_H = \{ \gamma_H^i \}_{i=1}^{i_{max}}$. W.l.o.g., both sequences have the same length as the agent with the shorter path can be assumed to stay at their final location for remainder of the other agent's path execution. Let  $c(\tilde{\Gamma})$ be the sum of pairwise distances between successive waypoints in a path $\tilde{\Gamma}$ and let $\delta(\tilde{\Gamma}_1, \tilde{\Gamma}_2)$ be $\delta(\tilde{\Gamma}_1, \tilde{\Gamma}_2) = max(d_{min}(\tilde{\Gamma}_1$, $\tilde{\Gamma}_2) - \sigma^{safe}, 0)$, where $\sigma^{safe}$ denotes the safety threshold and $d_{min}(\tilde{\Gamma}_1, \tilde{\Gamma}_2)$ is the minimum  Euclidean distance between $\tilde{\Gamma}_1$ and $\tilde{\Gamma}_2$:  $\emph{min}_{i=1,\ldots, i_{max}} \{d(\gamma_1^i, \gamma_2^i) \}$. Besides, let $c_C(a_c)$ be the cost of executing the communication action $a_c$, and $\eta_R$, $\eta_H $, $\eta_P$, and $\eta_C$ be the weights of the cost function. Using this notation, we define $J(n)$ as follows: 
\begin{align} 
   J(n) = \eta_Rc(\tilde{\Gamma}_R(n)) +\eta_Hc(\tilde{\Gamma}_H(n))+ \nonumber \\
   \eta_p1/\delta(\tilde{\Gamma}_R(n),\tilde{\Gamma}_H(n)) + \eta_Cc(a_c)   \label{eq:highcost}
\end{align}

In Alg.\,\ref{alg:highlevel}, at each planing iteration (lines 3-20), CP gets a library of motion plans $\Pi_R$ from MP. In lines 7-11, a branch of the tree is created for each $a_c$ and $\pi_R$. As explained in (\ref{eq:highcost}), the path-to-goal of $H$ and $R$ are required to compute a cost value for each branch. $\Gamma_H$ is thoroughly given by $T_H$, as mentioned in line \ref{alg:highlevel:th}. On the other hand, since a $\pi_R$ is likely a partial path, $T_R$ is utilized in line \ref{alg:highlevel:tr} to compute a completed path-to-goal for $R$ given $\pi_R$.

\begin{algorithm} 
\DontPrintSemicolon
\SetAlgoLined
 \KwInput{$\mathcal{P_{DC}}$}
 \KwOutput{$\Psi$}
 initialize: $\mathbf{b}_{0}$ and $\mathcal{S}^0$\;
 \While{$\textsc{Goal\_Test}(\mathcal{G}_R, \mathcal{S}_R)== \textsc{False}$}{ \label{alg:highlevel-while-start}
  $\Pi_R \leftarrow$ get the plans from the MP\;
  $\textsc{Min\_Cost} \leftarrow \infty$\;
  \For{$\pi_R  \in \Pi_R$} {
   \For{$a_c \in \mathcal{A}_c$} {
    $\omega_{k+1} \leftarrow O(a_c, \mathcal{S}^k$)\;
    $\mathbf{b}_{k+1} \leftarrow \textsc{Update}(\mathbf{b}_{k},\omega_{k+1}$)\;
    $\tilde{\Gamma}_H \leftarrow T_H(\mathcal{S}_H^k,\mathcal{G}_H,\mathbf{b}_{k+1})$\; \label{alg:highlevel:th}
    $\tilde{\Gamma}_R \leftarrow T_R(\mathcal{S}_R,\pi_R)$\; \label{alg:highlevel:tr}
    $c_{branch} \leftarrow J(\tilde{\Gamma}_R,\tilde{\Gamma}_H,a_c)$\;
    \If{$c_{branch} < \textsc{Min\_Cost}$}{
     $\textsc{Min\_Cost} \leftarrow c_{branch} $\\
     $\textsc{Best\_Action} \leftarrow \langle \pi_R,a_c\rangle$ }
   }
  } \label{alg:highlevel-while-end}
  $\textsc{Execute}(\textsc{Best\_Action})$ \;
  $\mathcal{S}^k \leftarrow$ $\mathcal{S}^{k+1}$ \;
  $\Psi.\textsc{append}(\textsc{Best\_Action})$
 }
 \caption{Communication Planner}
 \label{alg:highlevel}
\end{algorithm}

\begin{figure}[t]
      \centering
      \includegraphics[scale=0.42]{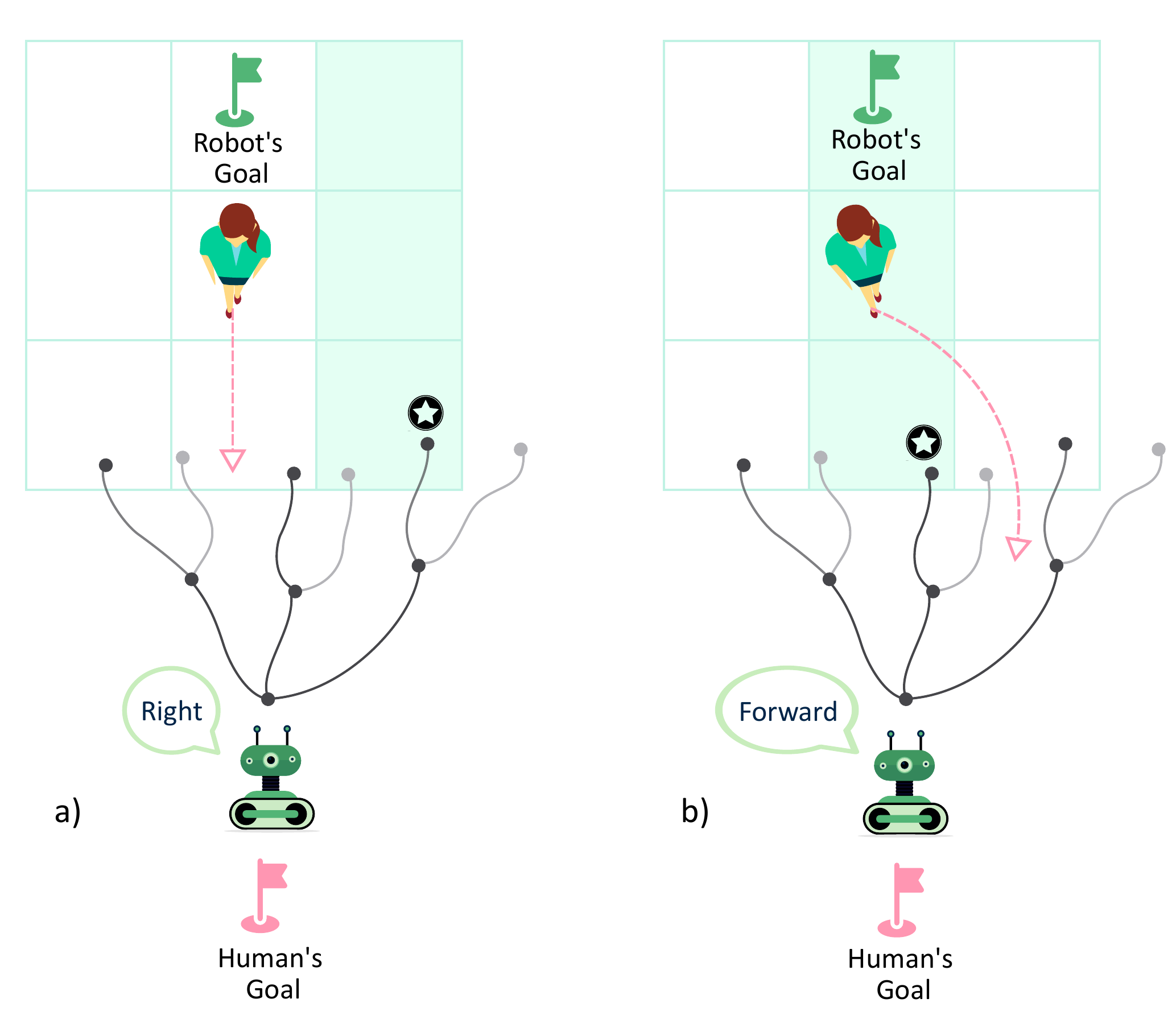}
      \caption{Two examples of the reasoning procedure of CP for a branch of the search tree.}
      \label{fig:example}
   \end{figure}

Fig.\,\ref{fig:example} exemplifies two branches of the CP search tree evaluated in lines 7-11 of Alg.\,\ref{alg:highlevel}. In each example, $H$ is shown at the center of its \emph{$\delta$-local neighborhood} visualized as a set of nine squares around her, where the colored squares stand for $\mathbf{b}_k$. Besides, $R$ is pictured at the bottom of each example with a partially expand RRT, where dark gray branches of RRT represented $\Pi_R$ selected by MP. In Fig.\,\ref{fig:example}(a), $R$ communicates $a_c =$``Right" and $\pi_R$ is the right branch of RRT, emphasized by a star, which makes $H$ believe that it will be in one of the squares on her left. In Fig.\,\ref{fig:example} (b), $R$ goes forward and communicates ``Forward" as well, which makes $H$ believe that $R$ will be in one of the middle squares. In scenario (b), $R$ takes a shorter path to goal but scenario (a) results in less conflicting paths for both $R$ and $H$. Thus, the more optimal branch will be determined based on the weights of $J$ in (\ref{eq:highcost}).     

\begin{Assumption}\label{assum: human-pred-highlevel}
The predicted trajectories $\Gamma_H$ given by $T_H$ in the discretized domain is an over-approximation of the predicted trajectories by $T_H$ in the continuous domain.
\end{Assumption}

\begin{Assumption}\label{assum: robot-pred-highlevel}
The discretized projection of $\Bar{\pi}_R$ on $\Gamma_R$ ($\pi_R$ in discretized domain) is an over-approximation of $\Bar{\pi}_R$ in the continuous domain.
\end{Assumption}

\begin{theorem}
Let $P_{DC}=\langle \mathcal{S}, s^0, \mathcal{A}, T, \mathcal{G}, O, J\rangle $ be a deliberative communication problem and let $\Psi^*=\langle(a^i_c,\pi^i_R)\rangle^{q}_{i=1}$ be its
solution  computed by Alg.\,\ref{alg:highlevel} using the cost function J in (\ref{eq:highcost}). Let $\Gamma_R$ be the discretized waypoints of $R$ in $\Psi^*$ defined as $\Gamma_R = \langle \pi_R^i\rangle_i $, and $\Gamma_H$ be a corresponding discretized waypoint sequence of a trajectory for $H$ predicted by $T_H$ and starting at $s^0$ with the goal $G_H$. If Assn. 1-3  hold, $\Gamma_R$ will either lie within $\Bar{\mathcal{S}}_R^{safe}$ or it will satisfy  $d_{min}(\Gamma_R, \Gamma_H) >\sigma^{safe}$.
\end{theorem}


\begin{proof}
Since $R$ has a null communication action that does not alter $H$'s belief, Alg.\,\ref{alg:highlevel} will always have a node reflecting the default behavior of CBF-TB-RRT with cost $<\infty$. In this case, Lemma\,\ref{lem:cbf safe} guarantees $R$'s trajectory not to leave $\Bar{\mathcal{S}}_R^{safe}$.
If Assn.\,\ref{assum: human-pred-highlevel} and\,\ref{assum: robot-pred-highlevel} hold, and if Alg.\,\ref{alg:highlevel} selects a node other than the default CBF-TB-RRT behavior, the min distance will be at least $\sigma^{safe}$, otherwise $\forall \, \eta_P > 0$, $J$ would be $\infty$ and the default CBF-TB-RRT behavior will be selected.
\end{proof}

\section {Empirical Evaluation}
We conducted extensive experiments in various simulation environments to evaluate the proposed method. These experiments 1) draw a comparison between the proposed method and the baseline method CBF-TB-RRT, and 2) illustrate the performance of the proposed method in deadlock situations. 

\subsection {Implementation}
\subsubsection{CBF-TB-RRT Design}
In our implementation, we consider the nonholonomic unicycle model for $R$ dynamics as 
\begin{align}\label{rot-model}
    \dot{\mathbf{s}}_r=\mathbf{g}_r(\mathbf{s}_r)\mathbf{a}_r ={\small\begin{bmatrix}
    \cos(\theta_r) & 0\\
    \sin(\theta_r) & 0\\
     0 & 1
    \end{bmatrix}}\mathbf{a}_r.
\end{align}
where states are $\mathbf{s}_r\!=\![x_r,y_r,\theta_r ]^T\!\in\!\mathcal{S}_R\!\subseteq \mathbb{R}^2\!\times\![-\pi,\pi)$ and control inputs are $\mathbf{a}_r=[v_r,\omega_r]^T\in \mathcal{A}_R\subseteq \mathbb{R}^2$. The parameters $x_r$, $y_r$, $\theta_r$ denote the longitudinal and lateral positions of $R$ and heading angle, respectively. The controls $v_r$ and $\omega_r$ also represent the linear and angular velocities of $R$, respectively.
Moreover, the goal set $\mathcal{S}_{g}\subset\mathcal{S}_R$ of $R$ can describe a set of position states in $\mathbb{R}^2$ as follows
\begin{align}\label{G-set2}
    \mathcal{S}_g = \big\{\mathbf{s}_r \in \mathcal{S}_R\;|\; \big\lVert [x_r,y_r]^T-\mathbf{s}_g\big\rVert^2_2 - r_g^2 \leq 0 \big\},   
\end{align}
where $\lVert\cdot\rVert_2$ denotes the Euclidean norm, $\mathbf{s}_g=[x_g,y_g]^T$ is the center, and $r_g$ is the radius of the goal set. \ 

While expanding the RRT tree, the following cost $c_i$ is assigned to each vertex $\nu_i\in\mathcal{V}$ for $i=0,1,\cdots,\lvert\mathcal{V}\rvert$,
\begin{align}
    c_i = w^G_dc^G_d+w^H_dc^H_d+w_gc_g+w_tc_t,
\end{align}
where $c^G_d$ is the Euclidean distance between vertex $i$ and the goal point, $c^H_d$ is the Euclidean distance between vertex $i$ and $H$, $c_h$ is the heading cost, and $c_t$ is the trap cost. The heading cost $c_g$ calculates the angular difference between the sampled vertex heading and the heading toward goal. To calculate the trap cost $c_t$, the algorithm checks the waypoints of a discretized direct straight line from the sampled vertex to the goal point. The trap cost $c_t$ is then the number of waypoints lied within the occupied regions. Readers are referred to \cite{majd2021safe} for further details on CBF-TB-RRT tree expansion. $w^G_d$, $w^H_d$, $w_g$, and $w_t$ are weight terms.\

   \begin{figure}[h!]
      \centering
      \includegraphics[scale=0.45]{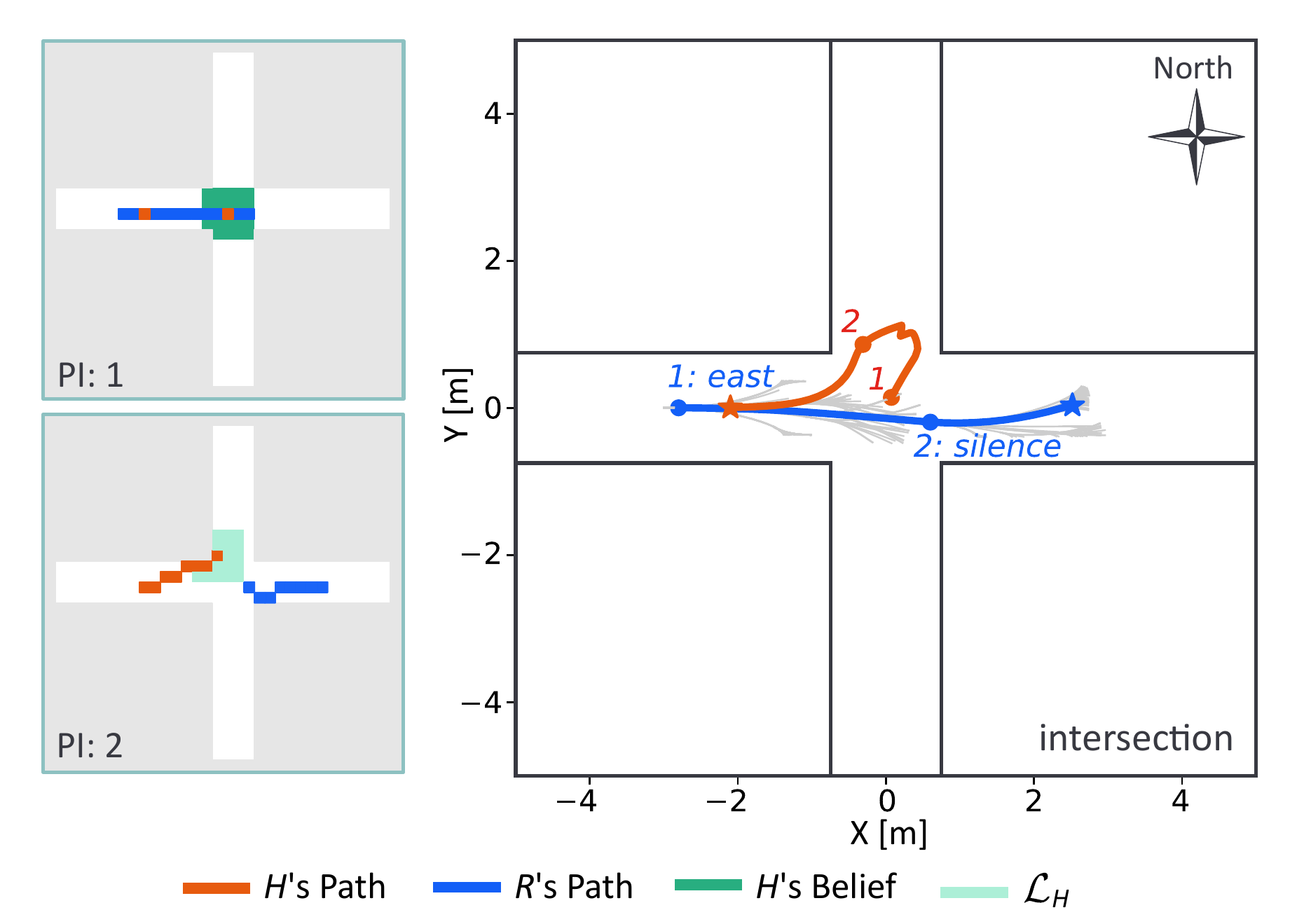}
      \caption{An example of a potential deadlock in confined environments.}
      \label{fig:deadlock}
   \end{figure}

\subsubsection{Human Movement Model}
We assumed $H$'s movement is described by a deterministic kinematic motion transition function ($T_H$) and we used the Dynamic Window Approach (DWA) in MP, proposed in \cite{fox1997dynamic}, to predict $H$'s shortest trajectory to the goal for a finite time horizon. Since DWA is a deterministic prediction method, we assumed an $\varepsilon$ bound around the human's predicted trajectory following Assn. \ref{assum: human-pred} to derive the CBF safety constraints. Given the human's predicted trajectory $\mathbf{s}_h$, we define the safe set $\mathcal{S}_R^{safe}\subseteq \mathcal{S}_R$ as $\mathcal{S}_R^{safe} = \big\{ \mathbf{s}_r \in \mathcal{S}_R, \mathbf{s}_h \in \mathcal{S}_H~\lvert ~B(\mathbf{s}_r,\mathbf{s}_h)\geq 0\big\}$, where $B(\mathbf{s}_r)$ is a continuously differentiable safety measure defined as 
\begin{align}\label{eq: safe-measure}
    B(\mathbf{s}_r,\mathbf{s}_h) = \lVert[x_r,y_r]^T - \mathbf{s}_h\rVert_2^2 - (\varepsilon + r_h + r_r)^2,
\end{align}
$r_h$, and $r_r$ are the radii of human and robot, respectively. The safety measure $B(\mathbf{s}_r)$ is employed as a CBF to impose the safety constraint (\ref{eq: CBF}) on the control input $\mathbf{a}_r$ in a Quadratic Program (QP) to generate safe plans $\pi_R$
\cite{majd2021safe}.\

As illustrated in Sec. \ref{sec:overview}, CP also utilizes $T_H$ to predict a trajectory-to-goal for $H$ for each branch of the search tree. Besides, in contrast to the requirements of the motion planning module, $H$ movement prediction must be provided for the whole horizon in communication planning module. Therefore, for the sake of computational efficiency, CP utilizes another $H$ movement model rather than DWA. CP considers a grid-based abstraction of the environment and utilizes A* search algorithm to predict a path-to-goal for $H$. In general this abstraction could be derived using methods for automatically predicting reliable state and action abstractions such as \cite{shah2022using}.
\begin{Assumption}
Predictions drawn from A* and DWA approaches complied with the Assn. \ref{assum: human-pred-highlevel} in all our experiments.
\end{Assumption}
\subsubsection{Human Motion Execution Model}
 We utilized the Social Forces model \cite{helbing_social_1995} to simulate the human movement, as it is very fast, scalable, and yet describes observed pedestrian behaviors  realistically. We modeled $H$ and $R$ both as pedestrians. To mimic $H$’s reactivity to $R$’s communication action $a_c$, the model creates multiple virtual agents moving from $R$’s current position to all $x$-$y$ projections of discretized zones $l_i\in l_H$ in the CP's belief model for which $b_i = 1$. If $\mathbf{b}_k =  \varnothing $, $R$’s goal is computed as a linear projection from its current position based on its current velocity, i.e. $H$ makes no assumptions over $R$’s future trajectory. Thus, in our experiments, the models used by $H$ are different from the model $H$ used by $R$, which is likely in real-world setting.

\subsection {Experimental Setup}
\begin{figure}[t]
      \centering
      \includegraphics[scale=0.5]{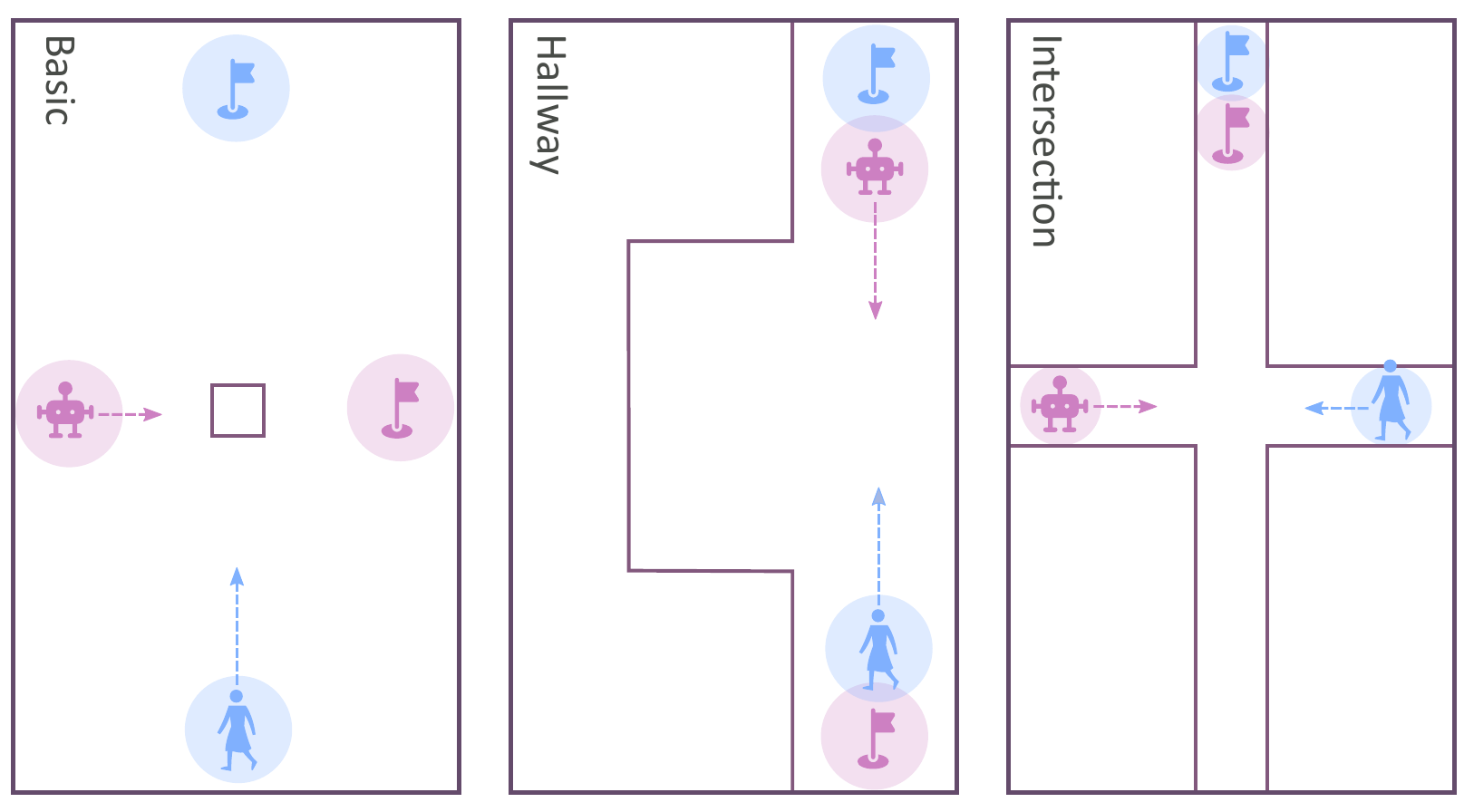}
      \caption{Schematic illustration of diversified test environments that capture various conflicting situation.}
      \label{fig:test_envs}
   \end{figure}

\emph{Test environments:} Fig. \ref{fig:test_envs} the environments used in our experiments. The basic floor map exemplifies spacious environments, while the hallway and intersection floor maps model more restricted and confined environments. 

\begin{table*}[h!]
\centering
\begin{threeparttable}
\caption{\label{tab:results}Comparison with CBF-TB-RRT.}
\begin{tabular}{|c|cccc|cccc|}
\hline 
\multirow{2}{*}{\diagbox[width=7em]{Maps}{Measures}} 
    & \multicolumn{4}{c|}{Our approach}  
    & \multicolumn{4}{c|}{CBF-TB-RRT}\\ 
    & {$R$ cost-to-goal} & {$H$ cost-to-goal} & {$PI$} & {$PC$} & {$R$ cost-to-goal} & {$H$ cost-to-goal} & {$PI$} & {$PC$}  \\
 \hline \hline
Basic 
    & 5.65\textendash5.68  & 7.33\textendash7.52 & 2\textendash2 & 0.50\textendash0.53 
    & 5.51\textendash6.27 & 6.90\textendash6.99 & 46\textendash113 & 0.21\textendash0.57 \\  
Intersection 
    & 3.63\textendash3.88 & 6.10\textendash6.29 & 2\textendash2 & 0.22\textendash0.24 
    & 4.20\textendash4.24 & 5.76\textendash5.90 & 51\textendash98 & 0.34\textendash $\infty$ \\  
Hallway 
    & 10.12\textendash10.58 & 6.85\textendash7.39 & 4\textendash4 & 0.72\textendash0.89 
    & 10.27\textendash10.30 & 6.65\textendash6.77 & 121\textendash123 & $\infty$ \textendash $\infty$ \\ 
    \hline

\end{tabular}
\begin{tablenotes}
      \small
      \item The results show the range of the measurements in 10 trials per map; PI: planning iterations; PC: proximity cost.
    \end{tablenotes}
    \end{threeparttable}
\end{table*}

   \begin{figure*}[h!]
      \centering
      \includegraphics[scale=0.33]{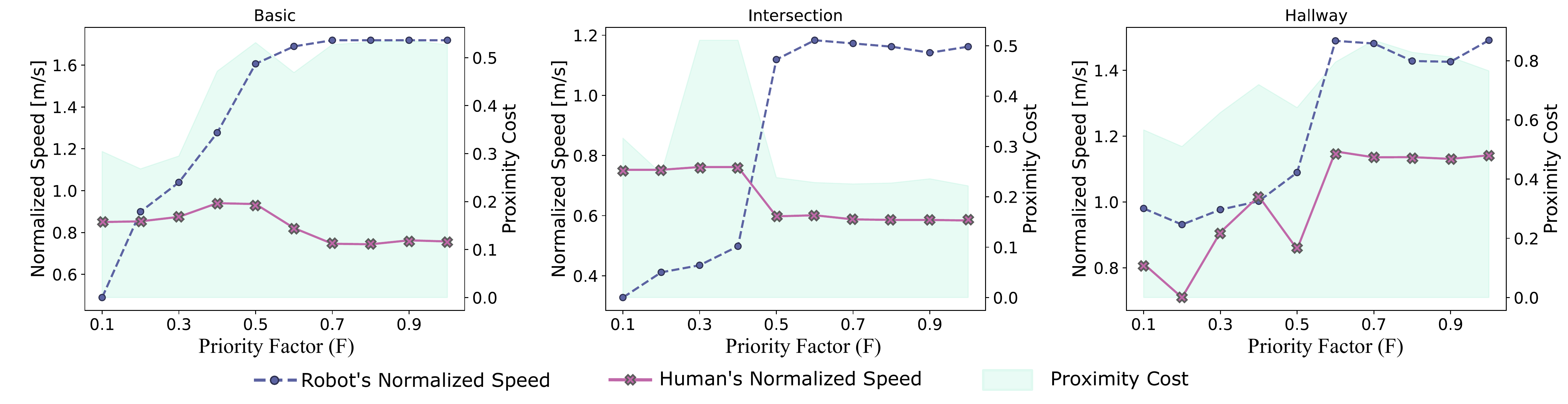}
      \caption{Flexible prioritization of $H$ and $R$ in different test environments, where $F=1$ prioritizes the robot.}
      \label{fig:weights}
   \end{figure*}

\noindent \emph{Measurements:}
Aside from cost-to-goal of $R$ and $H$, there are four more quantitative measures to evaluate the performance and effectiveness of the proposed method:
\begin {itemize}
    \item $R$'s normalized speed (RNS):  $RNS = \nicefrac{c_R^*}{time_R^{actual}}$ measures $R$'s normalized mean speed from $\mathbf{s}^0_r$ to $\mathcal{G}_R$, where $c_R^*$ and $time_R^{actual}$ denote the optimal cost-to-goal of $R$ and $R$'s actual travel time respectively. 
    \item $H$'s normalized speed (HNS):  $HNS = \nicefrac{c_H^*}{time_H^{actual}}$ measures $H$'s normalized average speed from $\mathbf{s}^0_h$ to $\mathcal{G}_H$, where $c_H^*$ and $time_H^{actual}$ denote the optimal cost-to-goal of $H$ and $H$'s actual travel time respectively.
    \item Planning iterations (PI): PI denotes the number of iterations of lines\,\ref{alg:highlevel-while-start} to\,\ref{alg:highlevel-while-end} in Alg.\,\ref{alg:highlevel}.
    \item Proximity cost (PC): PC measures the closeness of $R$ and $H$ during an experiments. Let $\Gamma_R = \{\gamma_R^i\}_{i=1}^{i_{max}}$ be $R$'s discretized trajectories given by a solution $\Psi$ and $\Gamma_H = \{\gamma_H^i\}_{i=1}^{i_{max}}$ be the corresponding discretized waypoint sequence of an actual trajectory for $H$. We defined PC using (\ref{eq: safe-measure}) as follows.
    \begin{align}
        Z = & \{ \zeta_i | \: \zeta_i = B(\gamma_R^i, \gamma_H^i) < thresh \}_{i=1}^{i_{max}} \\
        PC = &
            \begin{cases}
                \infty & \text{if $\exists \zeta_i \in Z, \: \zeta_i < 0$}\\
                \nicefrac{1}{\sum_{i=1}^{i_{max}}\zeta_i} & \text{otherwise}\\
            \end{cases},
    \end{align}\
\end {itemize}

\noindent \emph{Hypotheses:} throughout the experiments, we evaluate the following hypotheses 1) In confined environments, the chances of a deadlock are higher. Therefore, the effect of communication to avoid such deadlocks is more effective. 2) The proposed deliberative communication approach not only results in less conflicting social navigation, but also prevents deadlock situations where non-communicative approaches fail to find a solution. 3) By adjusting the weight vector of the cost function $J$, $H$ or $R$ can be prioritized. Accordingly, the non-prioritized agent is expected to have a decreased normalized average speed due to an increased cost-to-goal.

\subsection{Results}
\subsubsection{Comparison with CBF-RRT} \label{sec:compare}
In this section, we aim to demonstrate that the proposed method performs as optimally as CBF-TB-RRT, in terms of the traveled distances, while it reduces the conflict between $H$ and $R$. In Table \ref{tab:results}, the results are presented as the range of 10 experiments the experiments for each test environments of Fig. \ref{fig:test_envs}, where $\eta_R=1.5,  \eta_H=0.25, \eta_P = 3$, $\eta_C = 1$, and $A_c = \{ north, south,east, west \}$.

Our results show that $PC$ of the baseline drastically increases in more confined environments. E.g., $PC$ has a finite range in the basic environment since the room is spacious, while the $PC$ range is infinity in the intersection environment where the floor map is  confined and only one agent can pass through a corridor at a time. The situation is even more severe in the hallway environment in which the baseline method results in an infinite $PC$ for all 10 experiments. These observations validate 
Hypothesis 1. In contrast, the proposed method handles conflicting situations of the intersection and hallway environments effectively. 
The $PC$ values of our method in all environments are dramatically lower compared to the baseline method, while cost-to-goal of $R$ and $H$ do not increase noticeably. 

Moreover, employing the proposed method eliminates the necessity for frequent re-planning as $PI$ drops significantly compared to the experiments with the baseline method.

\subsubsection{Handling potential deadlocks}
According to \ref{sec:compare}, the proposed method is significantly more effective in reducing $PC$ in confined environments while maintaining the efficiency in terms of $c_R$ and $c_H$. This property is particularly imperative in preventing potential deadlocks in narrow passages, where a lower $PC$ implies less conflicting path for $H$ and $R$. Fig. \ref{fig:deadlock} demonstrates a pervasive case where lack of communication leads to a freezing situation. In this example, at the first planning iteration, $R$ transmits an ``east" signal, selected automatically by CP, to $H$ by which $H$ is informed about $R$'s plan before she enters the narrow corridor. As shown in Fig. \ref{fig:deadlock} (top left), this communication signal updates $H$'s belief about $R$'s next location adequately and impels $H$ to clear the passage. At the second planning iteration, $R$ has already passed through the intersection, so it remains silent and $H$'s belief indicates no collisions, as depicted in Fig. \ref{fig:deadlock} (bottom left).

In the same scenario, the baseline method performs ineffectively since $H$ enters the left corridor before $R$ departs it. When $H$ gets closer to $R$, there won't be enough room for the RRT to be expanded and a deadlock happens since the passage will be blocked for $R$ permanently. This analysis supports Hypothesis 2 regarding the capability of the proposed method to handle potential deadlocks.

\subsubsection{Flexible prioritization}
$H$ or $R$ can be prioritized flexibly by adjusting the weights of $J$. A parameter study on $\eta_R$ and $\eta_H$ reveals the way that each agent is favored in different social navigation scenarios, as shown in Fig. \ref{fig:weights}. In these experiments, the weights are adjusted 
as $\eta_R = F \eta_{const}$, and $\eta_H = (1-F) \eta_{const}$, where $F\in[0,1]$ denotes the priority factor ($R$ is fully prioritized for $F=1$), and $\eta_{const}=1.5$. In all three environments, prioritizing  $R$ increases $R$'s normalized speed significantly. Fig.\,\ref{fig:weights} shows that in the basic environment, $R$'s normalized speed increases by 2.7 times when $R$ is prioritized, compared to the case where $H$ is highly prioritized. Likewise, $H$ speeds up when she is prioritized in the basic and intersection environment. However, in the hallway environment, the whole $H$-$R$ interaction is relatively smoother and less conflicting when $R$ has a higher priority. Together, the present findings support Hypothesis 3. Furthermore, the results support the fact that the proposed method maintains a reasonably low $PC$ in all test environments not matter which agent is prioritized. In other words, the proposed method can be used to identify appropriate priorities for smooth social navigation.

\section {Conclusion}
This paper proposes a joint communication and motion planning framework that selects from an arbitrary input set of communication signals while computing the robot  motion  plans. The simulation results demonstrated that the presented framework avoids potential deadlocks in confined environments by leveraging explicit communications coupled with robot motion plans. We found that producing less conflicting trajectories for the robot in confined environments, which led to drastically lower proximity costs, indicates lower chances of a deadlock. We also observed that the proposed method does not degrade the robot's efficiency (in terms of traveled distances) compared to CBF-TB-RRT. In contrast, the non-communicative baseline method resulted in high proximity cost overall, which shows its incapability of generating viable solutions when extensive human-robot interaction is required. Furthermore, the proposed method can flexibly prioritize either the robot or the human while maintaining its effectiveness in handling potential deadlocks.

\bibliographystyle{ieeetr}
\bibliography{bibliography.bib}

\end{document}